\documentclass[letterpaper]{article} % DO NOT CHANGE THIS
\usepackage{aaai2026}  % DO NOT CHANGE THIS
\usepackage{times}  % DO NOT CHANGE THIS
\usepackage{helvet}  % DO NOT CHANGE THIS
\usepackage{courier}  % DO NOT CHANGE THIS
\usepackage[hyphens]{url}  % DO NOT CHANGE THIS
\usepackage{graphicx} % DO NOT CHANGE THIS
\usepackage{natbib}  % DO NOT CHANGE THIS AND DO NOT ADD ANY OPTIONS TO IT
\usepackage{caption} % DO NOT CHANGE THIS AND DO NOT ADD ANY OPTIONS TO IT

\usepackage{bm}
\usepackage{titletoc}
\usepackage{tcolorbox}
\usepackage{booktabs}
\usepackage{makecell}

\usepackage{algorithm}
\usepackage{algorithmic}
\usepackage{enumitem}  % 必须加载这个宏包
\usepackage[table]{xcolor}
\definecolor{Gray}{gray}{0.85}
\usepackage{newfloat}
\usepackage{listings}
\DeclareCaptionStyle{ruled}{labelfont=normalfont,labelsep=colon,strut=off} % DO NOT CHANGE THIS
\floatstyle{ruled}
\newfloat{listing}{tb}{lst}{}
\floatname{listing}{Listing}

\newtheorem{definition}{Definition}
\usepackage{amsmath}
\usepackage{amsfonts}
\usepackage{amssymb}
\usepackage{multirow}
\usepackage[T1]{fontenc} % 确保 Type 1 字体
\usepackage{pgfplots}
\pgfplotsset{compat=1.17}
\usepackage{pifont}
\pgfkeys{/pgf/text/.style={font=\rmfamily}} % 设置 Times Roman
\usepackage{subcaption}
\graphicspath{{figures/}} 

\definecolor{citecolor}{HTML}{2980b9}
\definecolor{linkcolor}{HTML}{c0392b}
\definecolor{lightblue}{RGB}{115, 192, 222}
\definecolor{lightgreen}{RGB}{145, 204, 117}
\definecolor{cornellred}{rgb}{0.7, 0.11, 0.11}
\definecolor{cadmiumgreen}{rgb}{0.0, 0.42, 0.24}
\definecolor{aliceblue}{rgb}{0.91, 0.94, 0.97}
\definecolor{darkblue}{rgb}{0.83, 0.89, 0.97}
\definecolor{Red7}{rgb}{0.941, 0.243, 0.243}
\definecolor{Green7}{RGB}{55, 178, 77}
\definecolor{Blue9}{rgb}{0.098,0.3,0.9}

\title{AsFT: Anchoring Safety During LLM Fine-Tuning \\ Within Narrow Safety Basin}

\author{
  Shuo Yang\textsuperscript{\rm 1}\thanks{Equal contribution},
  Qihui Zhang\textsuperscript{\rm 1}\footnotemark[1],
  Yuyang Liu\textsuperscript{\rm 1}\thanks{Corresponding author},
  Xiaojun Jia\textsuperscript{\rm 3},\\
  Kun-Peng Ning\textsuperscript{\rm 1},
  Jia-Yu Yao\textsuperscript{\rm 1},
  Jigang Wang\textsuperscript{\rm 4},
  Hailiang Dai\textsuperscript{\rm 4},
  Yibing Song\textsuperscript{\rm 5},
  Li Yuan\textsuperscript{\rm 1,2}\footnotemark[2]
}

\affiliations{
  \textsuperscript{\rm 1}Peking University, Shenzhen Graduate School\\
  \textsuperscript{\rm 2}Peng Cheng Laboratory\\
  \textsuperscript{\rm 3}Nanyang Technological University (NTU)\\
  \textsuperscript{\rm 4}ZTE Corporation\\
  \textsuperscript{\rm 5}Independent Researcher\\
  shuo\_yang@stu.pku.edu.cn, qhzhang25@stu.pku.edu.cn, yuanli-ece@pku.edu.cn
}

\begin{document}

\maketitle

\begin{abstract}
Fine-tuning large language models (LLMs) improves performance but introduces critical safety vulnerabilities: even minimal harmful data can severely compromise safety measures. We observe that perturbations orthogonal to the alignment direction—defined by weight differences between aligned (safe) and unaligned models—rapidly compromise model safety. In contrast, updates along the alignment direction largely preserve it, revealing the parameter space as a "narrow safety basin". To address this, we propose \textbf{AsFT} (\textbf{A}nchoring \textbf{S}afety in \textbf{F}ine-\textbf{T}uning) to maintain safety by explicitly constraining update directions during fine-tuning. By penalizing updates orthogonal to the alignment direction, AsFT effectively constrains the model within the "narrow safety basin," thus preserving its inherent safety. Extensive experiments on multiple datasets and models show that AsFT reduces harmful behaviors by up to 7.60\%, improves task performance by 3.44\%, and consistently outperforms existing methods across multiple tasks.
\end{abstract}

\begin{links}
\link{Code}{https://github.com/PKU-YuanGroup/AsFT}
\end{links}

\section{Introduction}

The rapid advancement of large language models (LLMs) has led to their widespread adoption, where fine-tuning is essential to adapt these models to specific tasks and scenarios.
However, fine-tuning exposes critical safety vulnerabilities. 
Even small amounts of malicious or harmless data during fine-tuning can compromise the model’s safeguards, causing it to generate harmful outputs post-fine-tuning~\cite{huang2024booster, bianchi2023safety, qi2023fine}. This raises the urgent need for methods that balance task-specific utility with robust safety defenses~\cite{huang2024trustllm}.

\begin{figure}[ht]
    \centering
    \includegraphics[width=\linewidth]{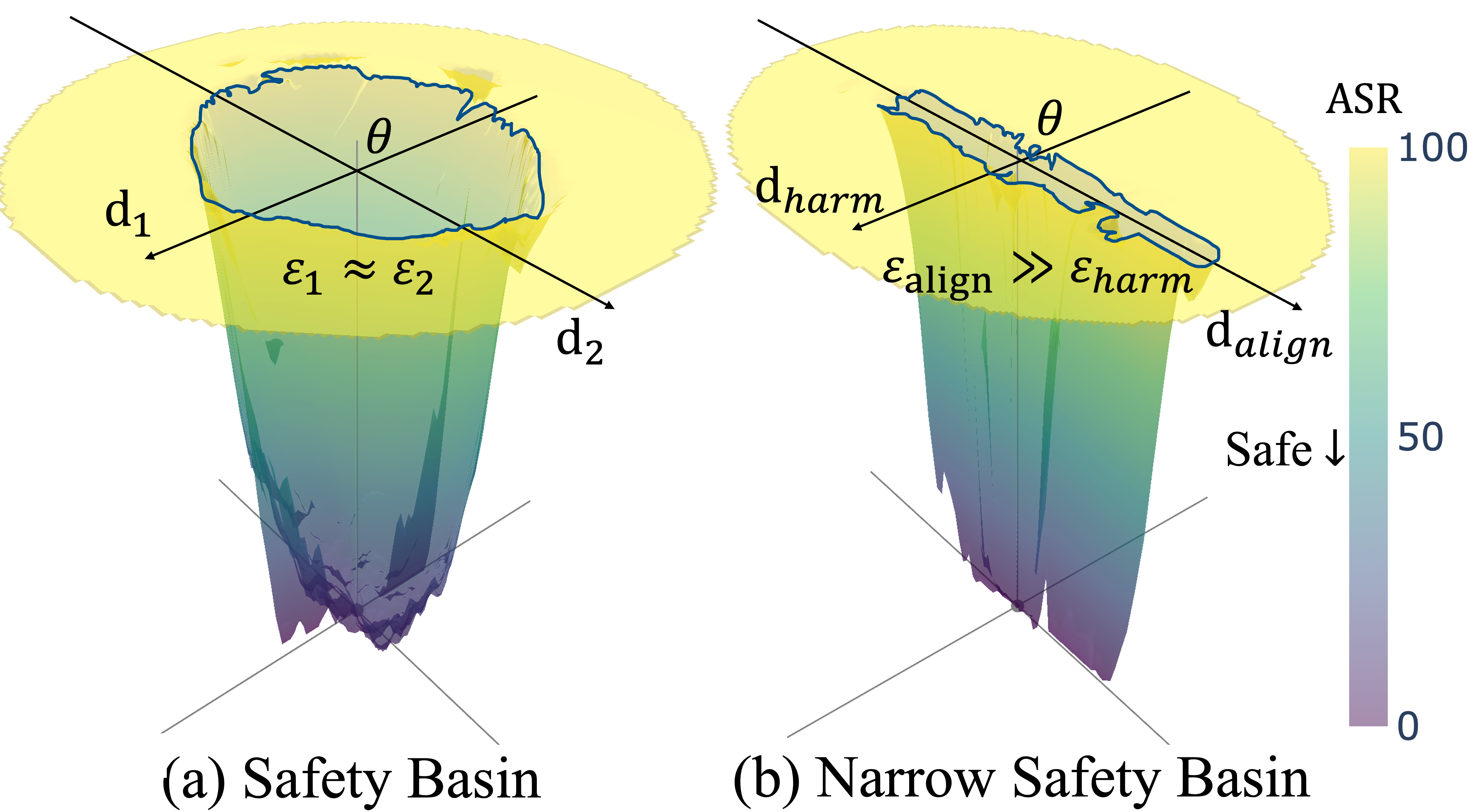}
    \caption{(a) The Safety Basin~\cite{peng2024navigating} shows a region where perturbations along $d_{\text{random}}$ preserve model safety, while safety sharply declines outside this area. (b) The Narrow Safety Basin demonstrates the asymmetry between $d_{\text{aligned}}$ and $d_{\text{harm}}$, where $d_{\text{aligned}}$ allows larger perturbations, while $d_{\text{harm}}$ causes sharp safety declines. In both subfigures, \textbf{lower values} indicate \textbf{higher safety}.}
    \label{fig1}
\end{figure}

Currently, there are various strategies for enhancing safety during LLM fine-tuning. While these strategies primarily rely on data-driven methods, they face a significant challenge: reliance on high-quality datasets, which are both costly and susceptible to bias~\cite{huang2024trustllm}.
Post-tuning methods like Safe LoRA~\cite{hsu2024safe} mitigate fine-tuning’s negative impact on model safety by discretizing and projecting LoRA weights into a safety-aligned subspace. 
However, they overlook layer continuity, as discrete projections can disrupt the consistency of learned features across layers. 
By focusing primarily on safety-related features, they neglect the performance-related characteristics brought by training data, degrading models' performance. 

{
\begin{figure*}[!t]
    \centering
    \includegraphics[width=0.9\textwidth]{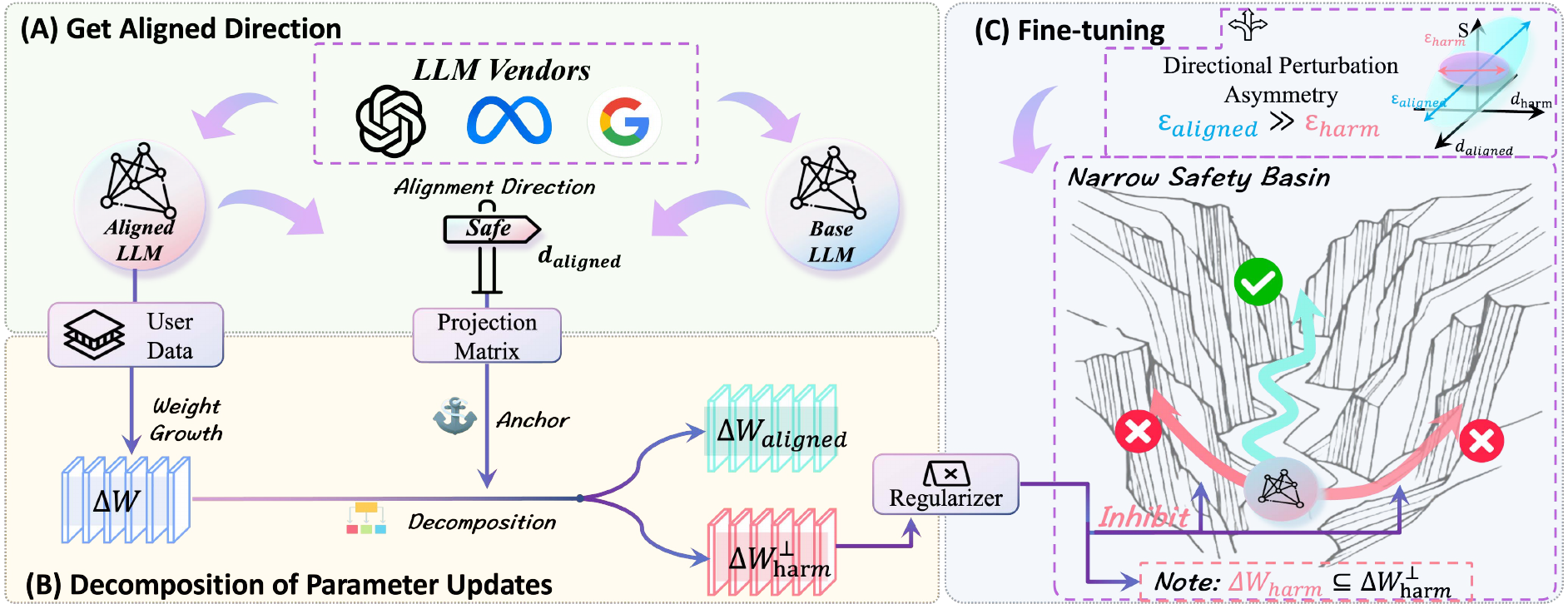}
    \caption{The proposed AsFT decomposes parameter updates into $d_{\text{aligned}}$ and $d^{\perp}_{\text{harm}}$, suppresses harmful updates along $d^{\perp}_{\text{harm}}$ by regularization and constrains updates within the narrow safety basin.}
    \label{fig_pipeline}
\end{figure*}
}

To address the limitations mentioned above, we aim to develop a data-free approach that leverages continuous optimization to enhance safety during fine-tuning. 
We observe that aligned models, developed under rigorous protocols, exhibit robust defenses against harmful inputs~\cite{qi2023fine, hsu2024safe}, whereas their unaligned counterparts (i.e., base models) lack such safeguards. This contrast inspires us to explore the latent information within the model parameter space.
The weight difference $\Delta \mathbf{W}$ between these two models encapsulates the alignment efforts undertaken by LLM vendors to enhance model safety. It not only reflects the core alignment process but also provides a critical direction for safety optimization~\cite{hsu2024safe, zhao2025identifying}. 

Given these observations, we hypothesize that the alignment direction can guide safety-preserving updates during fine-tuning and thus addresses the following question:

\begin{center}
    \textit{Can this weight difference serve as an anchor to guide safety-preserving updates?}
\end{center}

To investigate it, we explored the model's safety landscape~\cite{peng2024navigating} as shown in Fig.~\ref{fig1} and discovered a striking asymmetry: perturbations along the alignment direction ($d_{\text{aligned}}$, defined based on this weight difference $\Delta \mathbf{W}$) largely preserve model safety.
Conversely, direction orthogonal to it, which we term $d^{\perp}_{\text{harm}}$, is critically sensitive, where even small updates can trigger a sharp decline in safety.
This finding reframes the LLM parameter space as a ``narrow safety basin'' (Fig. 1(b)), a tight corridor where safety is maintained by moving along the alignment direction, while any deviation into the orthogonal space risks falling off a `safety cliff'.
% We define the alignment direction ($d_{\text{aligned}}$) based on this weight difference $\Delta \mathbf{W}$ and observe that perturbations along $d_{\text{aligned}}$ effectively preserve model’s safety. 
% Conversely, directions orthogonal to $d_{\text{aligned}}$ (denoted as $d_{\perp}$) are strongly correlated with harmful directions, where even small perturbations along $d_{\perp}$ can rapidly and significantly compromise the model’s safety.

% Conversely, directions orthogonal to $d_{\text{aligned}}$ are strongly correlated with harmful directions, and even small perturbations along these directions, which we define as $d^{\perp}_{\text{harm}}$, can rapidly and significantly compromise the model’s safety.

% This conceptualization frames the LLM parameter space as a ``narrow safety basin'' (as shown in Fig.~\ref{fig1}(b)), within which the model’s safety can be preserved by guiding updates along the constrained region defined by $d_{\text{aligned}}$.

% To navigate this treacherous landscape, we propose \textbf{SECURE} (\textbf{S}afety \textbf{E}nforcement \textbf{C}onstraint \textbf{U}sing \textbf{Re}gularized Orthogonality), a method (as shown in Fig.~\ref{fig_pipeline}) designed to keep the model firmly within the ``narrow safety basin'' during fine-tuning.  Instead of directly promoting updates along $d_{\text{aligned}}$, SECURE takes a more robust approach: penalizing any updates in the orthogonal direction $d^{\perp}_{\text{harm}}$ through a simple yet effective regularizer.

To navigate this treacherous landscape, we propose \textbf{AsFT} (\textbf{A}nchoring \textbf{S}afety in \textbf{F}ine-\textbf{T}uning), a method (Fig.~\ref{fig_pipeline}) that maintains models within the ``narrow safety basin" by penalizing parameter updates orthogonal to the alignment direction $d_{\text{aligned}}$. AsFT effectively prevents the model from straying into harmful regions of the parameter space, thus preserving its inherent safety while achieving strong task performance. Extensive experiment (across 8 datasets and 4 models) demonstrate that AsFT reduces harmful scores by up to 17.44$\%$ compared to SFT and achieves superior downstream performance. Our main contributions include:

% By doing so, SECURE prevents the model from straying into harmful regions of the parameter space, thus preserving its inherent safety while achieving strong task performance. Experimental results demonstrate that SECURE reduces harmful scores by up to 17.44$\%$ compared to SFT, while delivering superior performance on a variety of downstream tasks. In summary, our contributions are as follows:
% While the exact harmful direction is generally inaccessible, we use $d^{\perp}_{\text{harm}}$, derived from $d_{\text{aligned}}$, as a proxy to approximate and suppress harmful parameter updates. This is achieved by introducing a regularizer into the training objective, which explicitly constrains updates along $d^{\perp}_{\text{harm}}$ to guide them within the ``narrow safety basin'', effectively preserving the safety of the fine-tuned model while maintaining strong task-specific performance. 

\begin{itemize}[nolistsep, leftmargin=*]
    \item We observe that the alignment direction $d_{\text{aligned}}$ can serve as a safety anchor and that its orthogonal counterpart $d^{\perp}_{\text{harm}}$ closely aligns with the harmful direction, framing the LLM safety landscape as a ``narrow safety basin''.
    \item We propose \textbf{AsFT}, which penalizes parameter updates along $d^{\perp}_{\text{harm}}$, enabling fine-tuning within the ``narrow safety basin'' to preserve alignment safety.
    \item We validate AsFT through extensive experiments across \textbf{8 datasets} and \textbf{4 models}, achieving the best balance between safety and downstream task performance.
\end{itemize}

\section{Related Works}
% Ensuring the safety of models is critical, as they are increasingly deployed in sensitive and high-stakes scenarios where their outputs can significantly impact individuals and society~\cite{}.
Safety alignment ensures that large language models (LLMs) generate outputs aligned with human values and ethics \cite{touvron2023llama, zou2023representation, gao2023scaling, liu2025continual,tang2022neursafe, gao2023special}. Key techniques include instruction fine-tuning, RLHF, DPO, and others~\cite{wei2021finetuned, rafailov2024direct,yang2025look,yang2024parameter}. However, these methods are vulnerable to small-scale fine-tuning attacks, where minimal harmful or neutral data can compromise model safety \cite{qi2023fine, yao2023llm}. To address this, defenses have been developed across three stages: alignment, fine-tuning, and post-tuning \cite{huang2024harmful}.

\textbf{Alignment Phase Defenses} enhance model robustness against harmful fine-tuning attacks during the alignment phase \cite{qi2024safety, zhao2025understanding, liu2024robustifying}. Techniques such as Vaccine \cite{huang2024vaccine} introduce latent perturbations in the parameter space to ensure aligned outputs under adversarial conditions. RepNoise \cite{rosatirepresentation} removes harmful representations to prevent their reconstruction. TAR \cite{tamirisa2024tamper} optimizes parameters to maintain high harmful loss post adversarial fine-tuning, while Booster \cite{huang2024booster} minimizes harmful loss degradation during simulated attacks.

\textbf{Fine-tuning Phase Defenses} enhance safety during training against harmful fine-tuning~\cite{mukhoti2023fine, wei2024assessing, li2025safety}. MLLR~\cite{du2024towards} identifies critical modules with modular robustness analysis and applies differential learning rates. SafeInstr~\cite{bianchi2023safety} uses safety-focused examples. Lisa~\cite{huang2024lazy} limits optimization drift through dual-state optimization and proximity constraints. BEA~\cite{wangbackdooralign} embeds hidden triggers to suppress harmful content, while Seal~\cite{shen2024seal} removes harmful samples with two-stage optimization. SAFT~\cite{choi2024safety} filters harmful data using subspace decomposition scoring.

\textbf{Post-tuning Phase Defenses} aim to restore model safety after harmful fine-tuning attacks \cite{ ICLR2025_8a3cb724, yiprobe}. 
Safe LoRA \cite{hsu2024safe} discretely projects parameters onto the safe direction after fine-tuning.
SOMF \cite{yi2024safety} integrates additional benign task knowledge and reuses essential safety parameters.
Antidote \cite{huang2024antidote} effectively prunes harmful parameters during the post-processing stage, and SafetyLock \cite{zhu2024locking} leverages extracted safety directions to actively intervene in attention head activations during inference.

\section{Methodology}

\subsection{Preliminaries: Safety Landscape and Basin}
\label{Preliminaries}

The Safety Landscape, introduced by ~\citet{peng2024navigating}, characterizes how LLMs' safety varies across their parameter space, evaluated using a monotonic function $S(\cdot)$, where lower values indicate higher safety, typically measured as the Attack Success Rate (ASR). Let $\theta$ denote model weights, $d$ the perturbation direction, and $\alpha$ the perturbation magnitude, with $\hat{d} = d/|d|$ as a normalized direction. For two orthogonal directions, the safety landscape is defined as:

\begin{equation}
    f(\alpha, \beta) = S(\theta + \alpha \hat{d}_1 + \beta \hat{d}_2).
\end{equation}

In this context, \citet{peng2024navigating} identified the concept of a Safety Basin (as shown in Fig.~\ref{fig1}(a)). Therefore, we formalize this concept as $\mathcal{B}(\theta; \epsilon_1, \epsilon_2)$, which refers to a localized region in the parameter space where the model's safety remains robust against bounded perturbations, within the limits defined by the maximum allowable perturbations $\epsilon_1$ and $\epsilon_2$:

\begin{definition}[Safety Basin]
    The Safety Basin, denoted as $\mathcal{B}(\theta; \epsilon_1, \epsilon_2)$, is formally defined as
    \[
    \resizebox{\linewidth}{!}{$
    \begin{aligned}
    \mathcal{B}(\theta; \epsilon_1, \epsilon_2) = \Big\{ (\alpha, \beta) \in \mathbb{R}^2 \,\Big|\, 
    & S(\theta + \alpha \hat{d}_1 + \beta \hat{d}_2) \leq S_{\text{threshold}}, \\
    & |\alpha| \leq \epsilon_1, \,
    |\beta| \leq \epsilon_2, \,
    \hat{d}_1, \hat{d}_2 \sim \text{random}
    \Big\}.
    \end{aligned}
    $}
    \]
\end{definition}

\subsection{Rethinking the Safety Basin}
\label{section: Rethinking on Safety Basin}

The Safety Basin concept offers a theoretical basis for safety robustness. However, these initial explorations often treat the parameter space as isotropic, assuming the perturbations in random directions are uniform. This raises a critical question: Does the parameter space truly exhibit uniform safety properties in all directions, especially concerning the direction created by the safety alignment process itself? We hypothesize that the alignment process imparts a significant anisotropic structure to this landscape.

\textbf{Analysis of Alignment Direction.}  
To investigate this anisotropy, we define the alignment direction as \(d_{\text{aligned}} = \theta_{\text{aligned}} - \theta_{\text{unaligned}}\), which reflects the essential transformations for safety in the alignment process. To assess its distinct role, we empirically examined its relationships with directions from harmful (\(d_{\text{harm}}\)), benign (\(d_{\text{benign}}\)), and random (\(d_{\text{random}}\)) updates. We fine-tuned Llama-2-7B with varying amounts of harmful and benign data, ranging from 10 to 500 samples across five datasets~\cite{sheshadri2024targeted, zou2023universal, ji2024beavertails, mazeika2024harmbench, li2023alpacaeval}. This process allowed us to derive \(d_{\text{harm}}\) and \(d_{\text{benign}}\), as well as generate \(d_{\text{random}}\).

\begin{table}[t]
\centering
\setlength{\tabcolsep}{12pt}
\begin{tabular}{c|cc}
\toprule[1.5pt]
Num. & Harmful & Random \\
\hline
10  & $5.95 \times 10^{-4}$ & $8.486 \times 10^{-3}$ \\
20  & $5.67 \times 10^{-4}$ & $8.481 \times 10^{-3}$ \\
50  & $5.96 \times 10^{-4}$ & $8.489 \times 10^{-3}$ \\
100 & $7.28 \times 10^{-4}$ & $8.491 \times 10^{-3}$ \\
200 & $6.87 \times 10^{-4}$ & $8.490 \times 10^{-3}$ \\
500 & $6.05 \times 10^{-4}$ & $8.489 \times 10^{-3}$ \\
\hline
\textbf{Average} & \textbf{$6.30 \times 10^{-4}$} & \textbf{$8.488 \times 10^{-3}$} \\
\bottomrule[1.5pt]
\end{tabular}
\caption{Cosine similarity between $d_{\text{align}}$ and each of $d_{\text{harm}}$ and $d_{\text{random}}$, evaluated for different sample numbers.}
\label{tab:cos_similarity_all}
\end{table}

As shown in Tab.~\ref{tab:cos_similarity_all}, we calculated the cosine similarities between these directions and \(d_{\text{aligned}}\). Notably, \(d_{\text{harm}}\) is nearly orthogonal to \(d_{\text{aligned}}\), with cosine similarity consistently close to zero, confirming their near orthogonality across all amounts of harmful data. To validate that this high orthogonality is not a random occurrence, we compared the alignment direction's similarity with \(d_{\text{random}}\). The cosine similarity between \(d_{\text{aligned}}\) and \(d_{\text{harm}}\) is \(10^{-4}\), significantly lower than with random (\(10^{-3}\)) directions (a difference of 1–2 orders of magnitude). 
This indicates that \( d_{\text{harm}} \) exhibits much stronger orthogonality, with \( d_{\text{harm}} \gg d_{\text{random}} \), further confirming that the alignment direction encodes significant safety features in the parameter space.
This empirical evidence strongly supports our hypothesis that the alignment process induces an anisotropic structure in the parameter space, with the harmful update direction primarily lying in the subspace orthogonal to the alignment direction. Thus, we define the harmful direction orthogonal to \(d_{\text{aligned}}\) as \(d^{\perp}_{\text{harm}}\).

\textbf{Anisotropy of Safety Landscape.} 
Fig.~\ref{fig1}(b) illustrates the safety landscape along \(d_{\text{aligned}}\) and \(d_{\text{harm}}\). Perturbation  ranges along \(d_{\text{aligned}}\) are substantial, allowing the model to maintain safety within this range. In contrast, perturbations along \(d_{\text{harm}}\) are limited, signifying rapid safety degradation. The asymmetry in allowable perturbation ranges (\(\epsilon_{\text{aligned}} \gg \epsilon_{\text{harm}}\)) confirms the anisotropy of the safety landscape. 
Based on these findings, we formally define the landscape as ``narrow safety basin":
\begin{definition}[Narrow Safety Basin]
    The Narrow Safety Basin, $\mathcal{B}_{\text{narrow}}(\theta; \epsilon_1, \epsilon_2)$, satisfies:
    \[
    \resizebox{\linewidth}{!}{$
    \begin{aligned}
    \mathcal{B}_{\text{narrow}}(\theta; \epsilon_1, \epsilon_2) = \Big\{ (\alpha, \beta) \in \mathbb{R}^2 \,\Big|\, 
    & S(\theta + \alpha \hat{d}_{\text{aligned}} + \beta \hat{d}_{\text{harm}}) \leq S_{\text{threshold}}, \\
    & |\alpha| \leq \epsilon_1, \,
    |\beta| \leq \epsilon_2, \,
    \epsilon_1 \gg \epsilon_2
    \Big\}.
    \end{aligned}
    $}
    \]
\end{definition}
where, $\epsilon_1 \gg \epsilon_2$ indicates that the allowable perturbation range along $d_{\text{aligned}}$ is much larger than  $d_{\text{harm}}$. 

\subsection{Proposed Framework: AsFT}
Building on the observation that models' parameter updates along the harmful orthogonal direction $d^{\perp}_{\text{harm}}$ significantly compromise the model’s safety, we propose \textbf{AsFT} (Anchoring Safety in Fine-Tuning), which utilizes $d_{\text{aligned}}$ as an anchor to constrain updates within the ``narrow safety basin".

\textbf{Key Idea.} Identifying the purely harmful update direction is challenging due to the variability in harmful data distributions and differences in model architectures. In contrast, the alignment direction $d_{\text{aligned}}$ is relatively more accessible. Therefore,, we approximate the harmful direction using its orthogonal complement, $d^{\perp}_{\text{harm}}$, which captures potential harmful subspaces. The pipeline in Fig.~\ref{fig_pipeline} outlines the key steps: 1) computing $d_{\text{aligned}}$ and 2) incorporating a regularization term to suppress updates along $d^{\perp}_{\text{harm}}$.

\textbf{Decomposition of Parameter Updates.}
To analyze parameter updates during fine-tuning, we decompose parameter updates $\Delta \mathbf{W}$ into components along the alignment direction $d_{\text{aligned}}$ and its orthogonal $d^{\perp}_{\text{harm}}$. This decomposition allows us to isolate updates that may contribute to harmful behaviors, achieved through projection matrices:
\begin{equation}
    \label{Equ: decomposition}
    \Delta \mathbf{W} = C_{\text{aligned}} \Delta \mathbf{W} + C^{\perp}_{\text{harm}} \Delta \mathbf{W},
\end{equation}
where $C_{\text{aligned}}$ projects parameter updates onto $d_{\text{aligned}}$ and its orthogonal component $C^{\perp}_{\text{harm}}$ accordingly projects updates onto the remaining orthogonal subspace as follows:
\begin{equation}
\begin{array}{c}
C_{\text{aligned}} = d_{\text{aligned}} \left(d_{\text{aligned}}^{T} d_{\text{aligned}}\right)^{-1} d_{\text{aligned}}^{T}, \\
C^{\perp}_{\text{harm}} = I - C_{\text{aligned}}.
\end{array}
\end{equation}

The term $C^{\perp}_{\text{harm}} \Delta \mathbf{W}$ precisely isolates the component of the parameter update orthogonal to the alignment direction.  As our findings indicate that this subspace is the primary source of safety degradation, our core strategy is to directly suppress this component by penalizing its $\ell_2$ norm.

\textbf{Training Objective.} To mitigate potentially harmful updates and enforce this safety constraint during fine-tuning, we introduce a regularization term that specifically penalizes updates deviating from the alignment direction. Thus, our total loss function is defined as:
\begin{equation}
\mathcal{L} = \mathcal{L}_{\text{task}} + \mathcal{L}_{\text{reg}} = \mathcal{L}_{\text{task}} + \lambda \|C^{\perp}_{\text{harm}} \mathbf\Delta {W}\|^2,
\end{equation}
where $\mathcal{L}_{\text{task}}$ represents the original task loss associated with the specific objective, and $\lambda$ controls the regularization strength. By constraining the magnitude of $C^{\perp}_{\text{harm}}\Delta\mathbf{W}$, the regularizer maintains the model’s alignment with safety guidelines while preserving task performance.

\section{Experiments}
\subsection{Experimental Setups}
\label{Experimental Setups}

\textbf{Datasets.} 
We use a total of \textbf{eight datasets}: four primary datasets—SST2~\cite{socher2013recursive}, AGNEWS~\cite{zhang2015character}, GSM8K~\cite{cobbe2021training}, and AlpacaEval~\cite{li2023alpacaeval}—for fine-tuning tasks, and four harmful datasets—Harmful~\cite{sheshadri2024targeted} (default setting), AdvBench~\cite{zou2023universal}, BeaveTails~\cite{ji2024beavertails}, and HarmBench~\cite{mazeika2024harmbench}—to simulate harmful fine-tuning attacks. We mix a proportion $p$ of unsafe (poison) data from the harmful datasets with $(1-p)$ benign data, represented by $n_{\text{samples}}$.

\textbf{Models.} 
We evaluate our method with \textbf{four models} including Llama-2-7B-Chat~\cite{touvron2023llama}, Llama-3-8B-Instruct~\cite{dubey2024llama}, Gemma-2-9B-It~\cite{team2024gemma}, and Qwen-2-7B-Instruct~\cite{yang2024Qwen2}. By default, we set $p = 0.1$ and $n = 1000$, using Llama-2-7B-Chat as the baseline model unless stated otherwise.

\textbf{Baselines.} We compare AsFT against \textbf{six baselines}, including SFT (the vanilla supervised fine-tuning), Lisa (base and aligned)~\cite{huang2024lazy}, SafeInstr~\cite{bianchi2023safety}, BEA~\cite{wangbackdooralign}, and Safe LoRA~\cite{hsu2024safe}.

\textbf{Evaluation Metrics.}
Following \citet{huang2024booster}, we evaluate performance using two key metrics:
\begin{itemize}[nolistsep, leftmargin=*]
    \item \textbf{Fine-tuning Accuracy (FA)}: The top-1 accuracy on the test sets of fine-tuning tasks.
    \item \textbf{Harmful Score (HS)}: The proportion of unsafe outputs when the model encounters unseen malicious instructions, as determined by the audit model in~\citet{ji2024beavertails} and ~\citet{dubey2024llama3herdmodels}.
\end{itemize}

\textbf{Training Details.}
We employ LoRA~\cite{hu2021lora} for efficient fine-tuning of LLMs (the decomposition shown in Eq.~\ref{Equ: decomposition} corresponds to the LoRA weights), with a rank of 8 across all experiments. The AdamW optimizer is used with a learning rate of $5 \times 10^{-5}$, training for 10 epochs with a batch size of 8. The regularization coefficient $\lambda$ is set to 1. \textbf{Additional analysis} of the hyperparameters $\lambda$ and the learning rate is provided in section \ref{sec: Hyper-Parameter Analysis}.  We also provide comprehensive results for \textbf{full parameter} fine-tuning in section ~\ref{Sec: Discussion}.

\subsection{Experimental Results}

\begin{table*}[!ht]
    \centering
    \setlength{\tabcolsep}{5pt}
  \resizebox{1\linewidth}{!}{
    \begin{tabular}{c|c cccc c|ccccc c}
    \toprule[1.5pt]
        Methods &   \multicolumn{6}{c}{ \textbf{Harmful Score}  $\downarrow$}& \multicolumn{6}{c}{\textbf{Finetune Accuracy}  $\uparrow$}\\
           \cmidrule(lr){2-7} \cmidrule(lr){8-13}
$(n=1000)$  & clean& $p=0.05$& $p=0.1$& $p=0.15$ & $p=0.2$ & Average& clean& $p=0.05$& $p=0.1$& $p=0.15$ & $p=0.2$  & Average\\
 \midrule
SFT        & 2.40 & 16.40 & 17.60 & 24.40 & 46.80 & 21.52 & 82.90 & 81.00 & 84.30 & 84.30 & \textbf{83.80} & 83.26 \\

Lisa-base  & 26.40 & 24.00 & 27.20 & 31.20 & 22.80 & 26.32 & 75.70 & 63.80 & 73.50 & 72.30 & 65.60 & 70.18 \\
Lisa-aligned & 2.40 & 12.80 & 16.80 & 20.40 & 20.00 & 14.48 & 82.40 & 76.90 & 81.80 & 82.00 & 76.60 & 79.94 \\
SafeInstr   & 1.60 & 15.60 & 16.80 & 25.60 & 21.20 & 16.16 & \textbf{83.90} & 81.90 & 84.30 & \textbf{85.40} & \textbf{83.80} & \textbf{83.86} \\

BEA         & 4.80 & 15.80 & 16.40 & 21.60 & 16.40 & 14.80 & 82.60 & 78.30 & \textbf{84.40} & 81.00 & 69.10 & 79.08 \\

Safe LoRA   & 2.40 & 1.60 & 5.60 & \textbf{4.20} & 20.00 & 6.76 & 82.90 & 78.60 & 81.20 & 82.20 & 80.00 & 80.98 \\

AsFT (Ours)    & \textbf{1.60} & \textbf{2.00} & \textbf{4.00} & 6.80 & \textbf{6.00} & \textbf{4.08} & 83.00 & \textbf{84.30} & 84.30 & 84.50 & 82.80 & 83.78 \\
\bottomrule[1.5pt]
  \end{tabular}
}
\caption{Performance under different harmful ratios in the default setting.}
\label{table: harmful ratio}
\end{table*}

\subsubsection{Robustness to Poison Ratio}
We evaluate the trade-off between model safety and fine-tuning performance under varying poison ratios, with results summarized in Tab.~\ref{table: harmful ratio}. Compared to SFT, AsFT significantly reduces the harmful score while improving downstream task accuracy. SafeInstr shows slightly higher accuracy (0.1$\%$), but its harmful score is nearly four times greater. Compared to Safe LoRA, AsFT achieves a 2.68$\%$ lower harmful score and 2.80$\%$ higher accuracy, likely due to Safe LoRA’s discrete projection disrupting consistency. Overall, AsFT achieves the best balance between safety and performance across all poison ratios on other datasets.

\begin{table*}[!ht]
\setlength{\tabcolsep}{1pt}

\centering
 \resizebox{1\linewidth}{!}{
    \begin{tabular}{c|c cccc c|ccccc c}
         \toprule[1.5pt]
Methods &   \multicolumn{6}{c}{ \textbf{Harmful Score}  $\downarrow$}  & \multicolumn{6}{c}{\textbf{Finetune Accuracy}  $\uparrow$} \\
         \cmidrule(lr){2-7} \cmidrule(lr){8-13}
       $(p=0.1)$  & $n=500$ & $n=1000$ &$ n=1500$ & $n=2000$ & $n=2500$ & Average & $n=500$ & $n=1000$ & $n=1500$ & $n=2000$ & $n=2500$ & Average\\
 \midrule
SFT        & 12.40 & 17.60 & 14.80 & 16.80 & 12.40 & 14.80 & 82.70 & 84.30 & \textbf{84.20} & 84.70 & 84.80 & 84.14 \\
Lisa-base    & 25.20 & 27.20 & 24.80 & 25.20 & 24.40 & 25.36 & 59.70 & 73.50 & 80.50 & 82.00 & 81.90 & 75.52 \\
Lisa-aligned & 5.60 & 16.80 & 19.60 & 22.00 & 24.80 & 17.76 & 78.90 & 81.80 & 83.90 & 84.40 & 84.70 & 82.74 \\
SafeInstr   & 14.80 & 16.80 & 10.80 & 15.40 & 15.60 & 14.68 & 80.40 & \textbf{84.40} & 83.90 & 84.00 & 83.90 & 83.32 \\
BEA         & 13.60 & 16.40 & 9.20 & 11.20 & 14.00 & 12.68 & 76.50 & \textbf{84.40} & 83.70 & 81.00 & 83.10 & 81.64 \\
Safe LoRA   & \textbf{2.80} & 5.60 & 5.20 & 8.40 & 8.80 & 6.16 & 81.50 & 81.20 & 80.70 & 82.30 & 81.60 & 81.46 \\

AsFT (Ours)    & 4.00 & \textbf{4.00} & \textbf{2.40} & \textbf{1.60} & \textbf{4.00} & \textbf{3.20} & \textbf{82.80} & 84.30 & 83.90 & \textbf{85.30} & \textbf{86.00} & \textbf{84.46} \\
   \bottomrule[1.5pt]
 \end{tabular}
}
\caption{Performance under different sample numbers in the default setting.}
\label{table: sample number}
\end{table*}

\subsubsection{Generalization to Fine-Tuning Sample Number}
We evaluate the robustness of the methods across different sample numbers, with results summarized in Tab.~\ref{table: sample number}. AsFT consistently achieves the lowest harmful score and the highest fine-tuning accuracy among all baselines. Compared to Safe LoRA, we reduce the harmful score by 2.96$\%$ and improve fine-tuning accuracy by 3.00$\%$. Compared to SafeInstr, AsFT lowers the harmful score by 11.48$\%$ while maintaining 1.14$\%$ higher accuracy. Results demonstrate the robustness of AsFT across varying sample sizes, with consistent conclusions for more complex tasks.

\begin{table*}[!ht]
    \centering
\setlength{\tabcolsep}{10pt}

  \resizebox{1\linewidth}{!}{
    \begin{tabular}{c|cc|cc|cc|cc|cc}
    \toprule[1.5pt]
        Methods & \multicolumn{2}{c}{\textbf{Harmful}} & \multicolumn{2}{c}{\textbf{AdvBench}} & \multicolumn{2}{c}{\textbf{BeaveTails}} & \multicolumn{2}{c}{\textbf{HarmBench}} & \multicolumn{2}{c}{\textbf{Average}} \\
        \cmidrule(lr){2-3} \cmidrule(lr){4-5} \cmidrule(lr){6-7} \cmidrule(lr){8-9} \cmidrule(lr){10-11}
         (AGNEWS) & HS $\downarrow$ & FA $\uparrow$ & HS $\downarrow$ & FA $\uparrow$ & HS $\downarrow$ & FA $\uparrow$ & HS $\downarrow$ & FA $\uparrow$ & HS $\downarrow$ & FA $\uparrow$ \\
        \midrule
        SFT & 17.60 & 84.30 & 11.20 & 83.90 & 37.20 & 84.90 & 5.20 & 82.70 & 17.80 & 83.95 \\
        Lisa-base & 17.20 & 73.50 & 7.60 & 83.90 & 30.80 & 83.10 & 4.60 & 82.70 & 15.05 & 80.80\\
        Lisa-aligned & 16.80 & 81.80 & 4.80 & 82.60 & 31.40 & 85.80 & 5.80 & \textbf{84.30} & 14.70 & 83.63\\
        SafeInstr & 16.80 & 84.30 & 4.40 & \textbf{84.40} & 21.60 & 83.20 & 2.40 & 83.20 & 11.30 & 83.78\\
        BEA & 16.40 & \textbf{84.40} & 16.00 & 83.50 & 36.80 & 84.20 & 14.00 & 84.00 & 20.80 & \textbf{84.02} \\
        Safe LoRA & 5.60 & 81.20 & 4.00 & 82.30 & 18.80 & 82.60 & \textbf{2.00} & 81.70 & 7.60 & 81.95 \\
        
        AsFT (Ours) & \textbf{4.00} & 84.30  & \textbf{1.60} & 83.70 & \textbf{14.40} & 82.90 & 2.40 & 83.40 & \textbf{6.70} & 83.58 \\
        \bottomrule[1.5pt]
    \end{tabular}
  }

\caption{Performance under different harmful datasets (Harmful \cite{sheshadri2024targeted}, AdvBench \cite{zou2023universal}, BeaveTails \cite{ji2024beavertails}, and HarmBench \cite{mazeika2024harmbench} datasets) in the default setting.}
  \label{dif_harm_dataset}

\end{table*}

\subsubsection{Robustness to Poison Datasets}
We assess method robustness across various harmful datasets. Tab.~\ref{dif_harm_dataset} shows that while BEA has the highest fine-tuning accuracy, it also has a high harmful score (HS). Safe LoRA achieves the lowest HS but suffers a significant performance drop. In contrast, our method, AsFT, balances competitive accuracy (average 83.78\%) with a low harmful score (average 6.70\%), demonstrating robustness to diverse harmful data.

\subsubsection{Generalization to Fine-Tuning Datasets}
The performance of AsFT across four fine-tuning datasets is summarized in Tab.~\ref{table: datasets}. AsFT achieves significant reductions in harmful scores (HS), with improvements of 42.00$\%$, 13.60$\%$, 41.60$\%$, and 17.20$\%$, while delivering the lowest average HS and highest accuracy among all baselines. These indicate the effectiveness and strong generalization potential of AsFT across diverse tasks. 

%————————————————————————————————————————————————
\begin{table*}[!ht]
    \centering

    \setlength{\intextsep}{0pt}
    \setlength{\tabcolsep}{10pt}
  \resizebox{\linewidth}{!}{
    \begin{tabular}{c|cc|cc|cc|cc|cc}
    \toprule[1.5pt]
        Methods & \multicolumn{2}{c}{\textbf{SST2}} & \multicolumn{2}{c}{\textbf{AGNEWS}} & \multicolumn{2}{c}{\textbf{GSM8K}} & \multicolumn{2}{c}{\textbf{AlpacaEval}} & \multicolumn{2}{c}{\textbf{Average}} \\
        \cmidrule(lr){2-3} \cmidrule(lr){4-5} \cmidrule(lr){6-7} \cmidrule(lr){8-9} \cmidrule(lr){10-11}
         (Llama-2-7B) & HS $\downarrow$ & FA $\uparrow$ & HS $\downarrow$ & FA $\uparrow$ & HS $\downarrow$ & FA $\uparrow$ & HS $\downarrow$ & FA $\uparrow$ & HS $\downarrow$ & FA $\uparrow$ \\
         \midrule
SFT        & 48.00 & 94.50 & 17.60 & 84.30 & 56.00 & 23.80 & 20.40 & 49.80 & 35.50 & 63.10 \\
Lisa-base        & 27.60 & \textbf{96.90} & 27.20 & 73.50 & 35.20 & 24.00 & 25.20 & 35.85 & 28.80 & 57.56 \\
Lisa-aligned        & \textbf{5.60} & 93.58 & 16.80 & 81.80 & 16.00 & 19.40 & 4.80 & 57.30 & 10.80 & 63.02 \\
SafeInstr   & 9.20  & 93.35 & 16.80 & 84.30 & 17.60 & 19.30 & 10.80 & 42.70 & 13.60 & 59.91 \\
BEA         & 7.20  & 91.63 & 16.40 & \textbf{84.40} & 38.80 & 21.00 & 6.80  & 52.40 & 17.05 & 62.36 \\
Safe LoRA   & 11.20 & 89.24 & 5.60  & 81.20 & 36.00 & 23.60 & 5.20  & 54.70 & 14.50 & 62.19 \\

AsFT (Ours)    & 6.00  & 93.32 & \textbf{4.00}  & 84.30 & \textbf{14.40} & \textbf{26.00} & \textbf{3.20}  & \textbf{58.90} & \textbf{6.90}  & \textbf{65.63} \\
\bottomrule[1.5pt]
 \end{tabular}
}
\caption{Performance of models trained on different fine-tuning datasets with Llama-2-7B.}
\label{table: datasets}
\end{table*}

\begin{table*}[!ht]
    \centering
\setlength{\tabcolsep}{10pt}
  \resizebox{1\linewidth}{!}{
    \begin{tabular}{c|cc|cc|cc|cc|cc}
    \toprule[1.5pt]
        Methods & \multicolumn{2}{c}{\textbf{Llama-2-7B}} & \multicolumn{2}{c}{\textbf{Llama-3-8B}} & \multicolumn{2}{c}{\textbf{Qwen-2-7B}} & \multicolumn{2}{c}{\textbf{Gemma-2-9B}} & \multicolumn{2}{c}{\textbf{Average}} \\
        \cmidrule(lr){2-3} \cmidrule(lr){4-5} \cmidrule(lr){6-7} \cmidrule(lr){8-9} \cmidrule(lr){10-11}
         (AGNEWS) & HS $\downarrow$ & FA $\uparrow$ & HS $\downarrow$ & FA $\uparrow$ & HS $\downarrow$ & FA $\uparrow$ & HS $\downarrow$ & FA $\uparrow$ & HS $\downarrow$ & FA $\uparrow$ \\
        \midrule
        SFT & 17.60 & \textbf{84.30} & 73.60 & 90.30 & 49.20 & \textbf{90.30} & 32.00 & \textbf{88.30} & 43.10 & \textbf{88.30} \\
        Lisa-base & 27.20 & 63.80 & 29.60 & 77.30 & 28.00 & 79.90 & 31.20 & 80.00 & 29.00 & 75.25\\
        Lisa-aligned   & 16.80 & 81.80 & 19.60 & 88.10 & 27.60 & 89.20 & 14.70 & 85.60 & 19.68 & 86.18 \\
        Safe LoRA & 5.60 & 81.20 & 26.40 & 87.80 & 8.40 & 85.50 & 8.40 & 84.70 & 12.20 & 84.8 \\
        SafeInstr & 16.80 & 84.40 & 18.80 & 89.00 & 7.20 & 83.30 & 7.60 & 84.70 & 12.60 & 85.35 \\
        BEA & 16.40 & 84.40 & 30.80 & 88.8 & 8.40 & 88.60 & 7.20 & 86.20 & 15.70 & 87.00 \\
        
        AsFT (Ours) & \textbf{4.00} & \textbf{84.30} & \textbf{15.20} & \textbf{92.30} & \textbf{5.20} & 87.90 & \textbf{6.00} & 86.60 & \textbf{7.60} & 87.78 \\
        \bottomrule[1.5pt]
    \end{tabular}
  }
\caption{Performance of different architectures evaluated on various metrics.}
\label{models}
\end{table*}

\subsubsection{Generalization to Models}
We evaluate methods across various architectures, as shown in Tab.~\ref{models}. AsFT consistently achieves the lowest harmful score (HS) and competitive accuracy, providing the best trade-off among baselines. It reduces HS by 36.00\% and improves accuracy by 1.00\% for models in the same architecture family (e.g., Llama-2 and Llama-3). AsFT also excels with other architectures like Qwen-2 and Gemma-2, maintaining an optimal balance between safety and performance, which is consistent in challenging tasks like \textbf{GSM8K}.

\subsection{Further Analysis of Narrow Safety Basin}
\begin{figure}[ht]
    \centering
  \includegraphics[width=\linewidth]{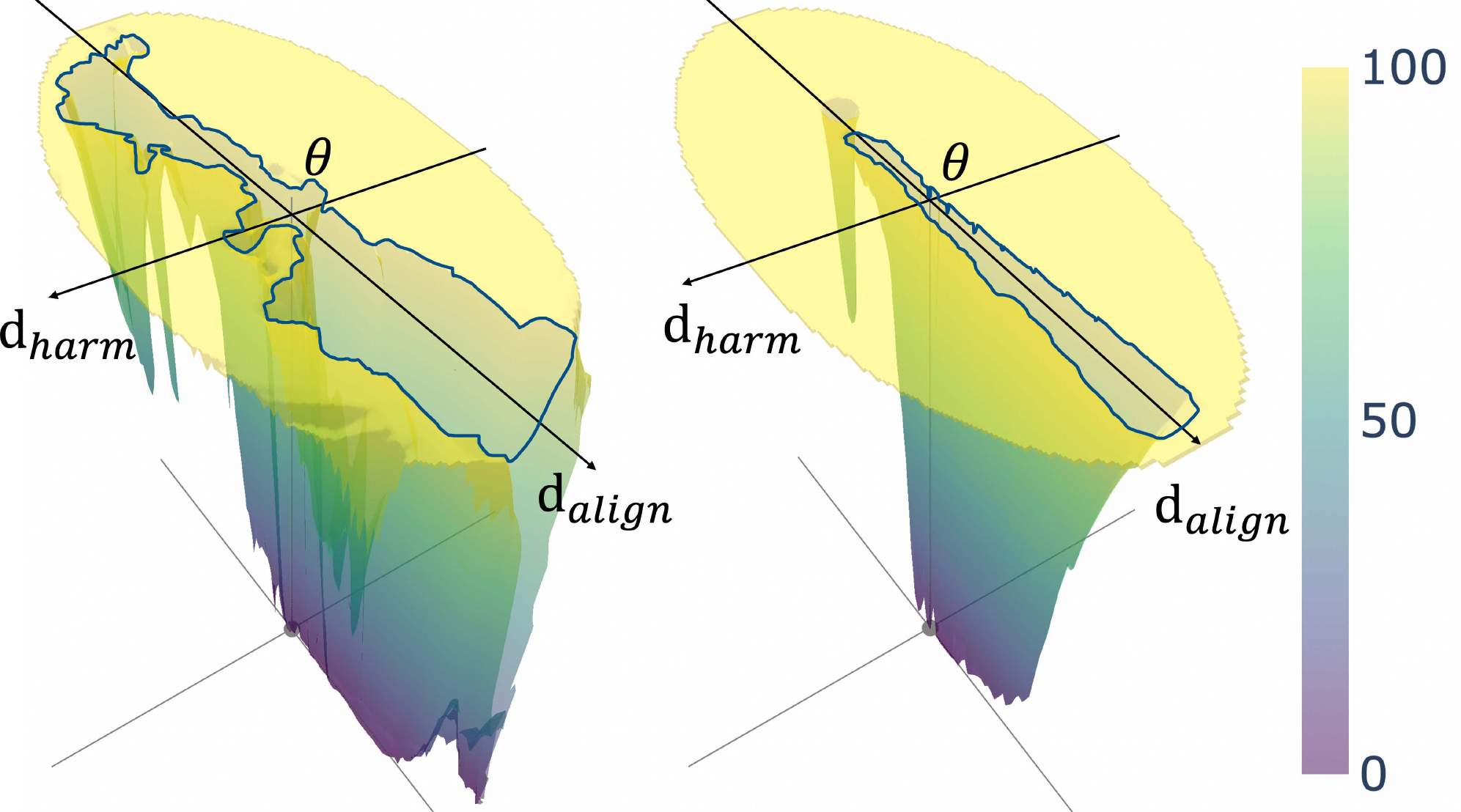}
  \caption{Safety landscape of Qwen-2-7B (left) and Gemma-2-9B (right) anchored along $d_{\text{aligned}}$.}
  \label{fig: landscape}
\end{figure}

To visualize the LLM safety landscape, we follow the methodology of~\citet{peng2024navigating}, anchoring our analysis on the alignment direction $d_{\text{aligned}}$ and sampling 20 directions. We plot the safety landscapes for Llama-2-7B (Tab.~\ref{fig1}(b)), Qwen-2-7B, and Gemma-2-9B (Tab.~\ref{fig: landscape}). Despite architectural differences, the visualizations reveal a consistent narrow safety basin, underscoring similarities across model architectures.

% \begin{table}[!ht]
% \centering
% % \setlength{\tabcolsep}{pt}
% \begin{tabular}{c | c | c}
% \toprule[1.5pt]
% % Models & \makecell[c]{Alignment direction \\ $d_{\text{aligned}}$} & \makecell[c]{Harmful direction\\ $d_{\text{harm}}$}   \\
% Models & $d_{\text{aligned}}$ & $d_{\text{harm}}$   \\
% \midrule
% Llama-2   & 0.1287  & 0.0099 \\
% Qwen-2   & 0.6594  & 0.0149 \\
% Gemma-2    & 0.3069  & 0.0046  \\
% \bottomrule[1.5pt]
% \end{tabular}

% \caption{EPL values for three models along $d_{\text{aligned}}$ and $d_{\text{harm}}$, which represent relative perturbation tolerance}
% \label{tab:EPL}
% \end{table}

\begin{table}[!ht]
\setlength{\tabcolsep}{12pt}
\centering
\begin{tabular}{c | c  c  c}
\toprule[1.5pt]
Models & Llama-2 & Qwen-2 & Gemma-2 \\
\midrule
$d_{\text{aligned}}$ & 0.1287 & 0.6594 & 0.3069 \\
$d_{\text{harm}}$    & 0.0099 & 0.0149 & 0.0046 \\
\bottomrule[1.5pt]
\end{tabular}
\caption{EPL values for three models along $d_{\text{aligned}}$ and $d_{\text{harm}}$, which represent relative perturbation tolerance.}
\label{tab:EPL}
\end{table}

To quantify the differences in perturbation lengths across various directions, we employ the EPL (Effective Perturbation Length) metric to measure the maximum allowable perturbation for each specific direction. It is defined as:

% \begin{equation}
%     \resizebox{1.0\linewidth}{!}{$\text{EPL} = \sup \left\{ |\alpha| \,  \middle| \, \mathcal{S}(\theta +\alpha d) \geq \tau, \,
%     \alpha \in \mathcal{U}(-a, a), \, d \in D \right\}$}
% \end{equation}
\begin{equation}
\begin{split}
    \text{EPL} ={}& \sup \left\{ |\alpha| \, \middle| \, \mathcal{S}(\theta +\alpha d) \geq \tau, \right. \\
                 & \qquad \left. \alpha \in \mathcal{U}(-a, a), \, d \in D \right\},
\end{split}
\end{equation}
where $\alpha$ is the perturbation magnitude, and $d$ is its direction. Tab.~\ref{tab:EPL} shows EPL values for three models along $d_{\text{aligned}}$ and $d_{\text{harm}}$ (the latter closely related to $d^{\perp}_{\text{harm}}$). Higher EPL values along $d_{\text{aligned}}$ indicate greater robustness to safety-preserving perturbations, while lower values along $d^{\perp}_{\text{harm}}$ reveal sensitivity to harmful directions. These results highlight the anisotropic nature of landscape and the importance of $d_{\text{aligned}}$ in guiding updates within the narrow safety basin.

\begin{figure*}[!t]
  \includegraphics[width=1\textwidth]{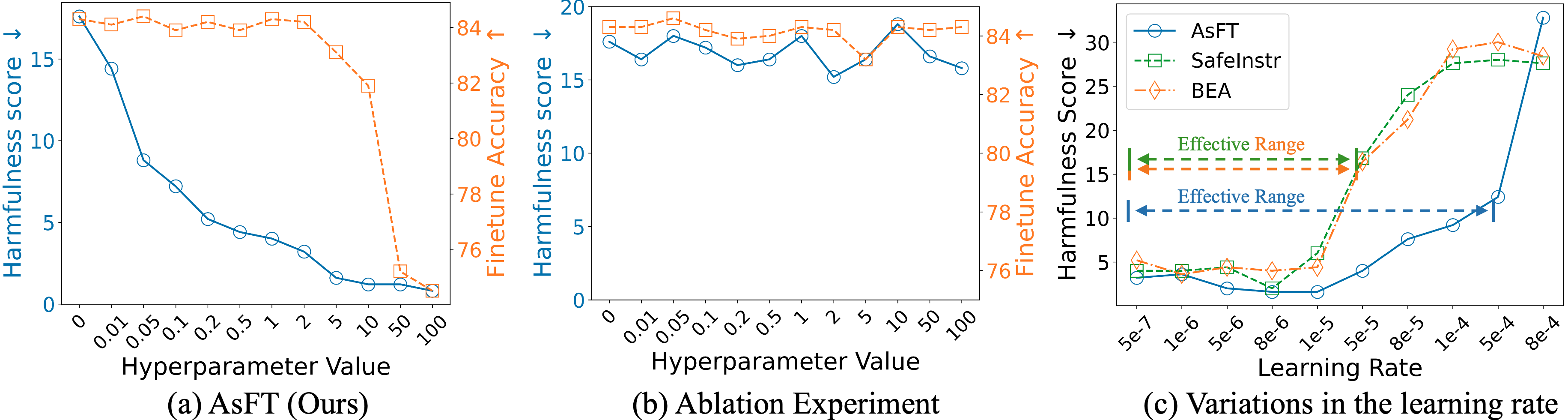}
  \caption{(a) Restricting updates along $d^{\perp}_{\text{harm}}$ (AsFT) significantly reduces harmful scores as $\lambda$ increases, while maintaining fine-tuning accuracy. (b) Restricting updates along $d_{\text{aligned}}$ results in consistently high harmful scores. (c) Comparison of robustness to learning rate variations shows that AsFT achieves a broader effective range compared to data-driven methods 
 (SafeInstr~\cite{bianchi2023safety} and BEA~\cite{wangbackdooralign}).}
  \label{fig: Ablation Experiment}
\end{figure*}

\begin{table*}[!t]
\setlength{\tabcolsep}{1pt}
\centering
  \resizebox{1\linewidth}{!}{
    \begin{tabular}{c|c cccc c|ccccc c}
    \toprule[1.5pt]
    Methods &   \multicolumn{6}{c}{ \textbf{Harmful Score}  $\downarrow$}& \multicolumn{6}{c}{\textbf{Finetune Accuracy}  $\uparrow$}\\
       \cmidrule(lr){2-7} \cmidrule(lr){8-13}
    (AGNEWS) & $n=500$ & $n=1000$ & $n=1500$ & $n=2000$ & $n=2500$ & Avg  & $n=500$ & $n=1000$ & $n=1500$ & $n=2000$ & $n=2500$ & Avg \\
    \midrule
    SFT                 & 12.40 & 17.60 & 14.80 & 16.80 & 12.40 & 14.80 & 82.70 & 84.30 & 84.20 & 84.70 & 84.80 & 84.14 \\
    
    AsFT$_\text{Alt}$ & \textbf{5.60}  & \textbf{9.60}  & \textbf{8.80}  & \textbf{12.80} & \textbf{8.40}  & \textbf{9.04}  & \textbf{83.00} & \textbf{84.00} & \textbf{83.80} & \textbf{85.30} & \textbf{85.80} & \textbf{84.38} \\
    \bottomrule[1.5pt]
\end{tabular}
}
\caption{The alternative AsFT$_\text{Alt}$ still significantly reduces harmful outputs while maintaining competitive task performance.}
\label{Alt-table: diff sample compare}
\end{table*}

\subsection{Hyper-Parameter Analysis and Ablation Experiments}
\label{sec: Hyper-Parameter Analysis}

\subsubsection{Robustness to Hyper-Parameter $\bm{\lambda}$}
Tab.~\ref{fig: Ablation Experiment} (a) shows that as $\lambda$ increases from 0 (SFT), the harmful score (HS) decreases while accuracy remains stable, until $\lambda > 10$ where accuracy drops. This suggests an optimal $\lambda$ range of 0.1 to 10. To further demonstrate robustness, we conducted additional experiments on diverse datasets. Across these datasets, AsFT consistently achieves a stable safety-performance trade-off within this broad two-order-of-magnitude range for $\lambda$. This indicates that our approach does not require meticulous hyperparameter tuning, as selecting $\lambda$ between 0.1 and 10 is generally sufficient to significantly reduce harmful outputs while preserving performance.

\subsubsection{Ablation Experiment}
The ablation results in Fig.\ref{fig: Ablation Experiment} evaluate the impact of constraining parameter updates along different directions. 
In Fig.\ref{fig: Ablation Experiment}(a), we restrict updates along the orthogonal direction $d^{\perp}_{\text{harm}}$, as in \textbf{AsFT} (updating along the narrow safety basin). This restriction leads to a clear reduction in harmful scores (HS) with increasing $\lambda$, demonstrating the effectiveness of AsFT in improving safety while maintaining accuracy. In contrast, Fig.\ref{fig: Ablation Experiment}(b) shows that restricting updates along the alignment direction $d_{\text{aligned}}$ (updating perpendicular to the narrow safety basin) does not result in a reduction of HS, which remain high across all $\lambda$ values. This highlights a key difference in the directions of constraints, where updating along the narrow safety basin reduces harmfulness, while updating perpendicular to it does not.

\subsubsection{Robustness to Learning Rate}
Fig.\ref{fig: Ablation Experiment} (c) compares the robustness of AsFT with data-driven defenses like SafeInstr and BEA under varying learning rates.
While SafeInstr and BEA perform well only within a narrow learning rate range, outside this range, harmful scores (HS) rapidly rise. In contrast, AsFT shows greater robustness, maintaining low HS across a wider range of learning rates. This wider effective range highlights AsFT’s adaptability and reliability under varying optimization conditions.

\section{Discussion}
\label{Sec: Discussion}

\textbf{Effectiveness in Full-Parameter Fine-Tuning.} The efficacy of AsFT is fundamentally rooted in the ``narrow safety basin'' phenomenon, an observed characteristic of the model's complete parameter landscape. This makes our method effective for both LoRA-based and full-parameter fine-tuning. 
When extended to full-parameter fine-tuning, AsFT consistently achieved superior results by reducing harmful scores while maintaining high fine-tuning accuracy. 
% When all methods were extended to full-parameter fine-tuning, AsFT consistently achieved superior results by reducing harmful scores while maintaining high fine-tuning accuracy. 

\textbf{Method Adaptability.}
Many mainstream open-source models, such as Qwen and Llama, typically provide both their aligned and base model weights. This common practice ensures that our method, which assumes their availability, is broadly applicable.
Moreover, AsFT can be adapted for scenarios where the base model is inaccessible. Specifically, harmful data can be used to identify harmful directions, and the fine-tuning process can then be guided by the orthogonal complement to these directions. As shown in Tab.~\ref{Alt-table: diff sample compare}, AsFT$_\text{Alt}$ significantly reduces harmful outputs while maintaining competitive task performance.

\textbf{Further Evaluation in Challenging Scenarios.}
We further evaluated the robustness and reliability of AsFT in more challenging and diverse scenarios. 
% Our further evaluation of AsFT's robustness and reliability in more challenging and diverse scenarios demonstrated that under high poison ratio settings, AsFT reduced the HS to 19.70, significantly outperforming SFT (63.50) and Safe LoRA (25.10), while maintaining competitive FA (81.73$\%$).
Specifically, we tested AsFT against two representative \textbf{jailbreak techniques}, LLM-DRA~\cite{liu2024making} and ArtPrompt~\cite{jiang2024artprompt}, and found that it maintained robust performance under adversarial conditions. Additionally, we increased the proportion of harmful data up to 60$\%$, showing that AsFT remained both safe and effective even in these \textbf{more difficult settings}. To further enhance the reliability of our harmfulness assessment, we incorporated Llama-Guard-3-8B~\cite{dubey2024llama3herdmodels} as an \textbf{additional safety evaluator}, with results from both evaluators closely aligned.

\section{Conclusion}
In this work, we address the safety vulnerabilities of large language models (LLMs) during fine-tuning by introducing \textbf{AsFT} (Anchoring Safety in Fine-Tuning), a method that anchors parameter updates within the safety-preserving alignment direction ($d_{\text{aligned}}$). By regularizing updates along the orthogonal direction ($d^{\perp}_{\text{harm}}$), AsFT reduces harmfulness while preserving task performance. Extensive experiments show that AsFT outperforms existing methods, achieving lower harmful score and higher accuracy, which emphasize the value of limiting updates within the narrow safety basin to ensure safety of LLMs.

\section*{Acknowledgements}
This work was supported by the China Postdoctoral Science Foundation under Grant Number BX20240013 and 2024M760113, the Natural Science Foundation of China (No. 62332002, 62425101), Shenzhen Science and Technology Program (KQTD20240729102051063), and ZTE$\&$PKU joint lab (No.IA20241211013).

\bibliography{bibliography/aaai2026}
% \input{sections/AAAI26_Checklist}

% \clearpage
% \input{appendix/supplementary}

\end{document}